\documentclass[conference]{IEEEtran}
\IEEEoverridecommandlockouts
\usepackage{cite}
\usepackage{amsmath,amssymb,amsfonts}
\usepackage{algorithmic}
\usepackage{graphicx}
\usepackage{textcomp}
\usepackage{xcolor}
\usepackage{booktabs}
\usepackage{float}
\usepackage{placeins}
\DeclareMathOperator*{\argmax}{arg\,max}
\def\BibTeX{{\rm B\kern-.05em{\sc i\kern-.025em b}\kern-.08em
    T\kern-.1667em\lower.7ex\hbox{E}\kern-.125emX}}
\newcommand{\secgap}{\vspace{0.55\baselineskip}}

\title{Safety Beyond the Interface: Detecting Harm via Latent States in Large Language Models}

\author{
\IEEEauthorblockN{Alizishaan Khatri}
\IEEEauthorblockA{\textit{Wrynx Inc.}\\
\texttt{research@wrynx.com}}
\and
\IEEEauthorblockN{Chiquita Prabhu}
\IEEEauthorblockA{\textit{Independent Researcher}}
\and
\IEEEauthorblockN{Omkar Neogi}
\IEEEauthorblockA{\textit{Independent Researcher}}
}

\begin{document}
\maketitle

\begin{abstract}
Autonomous systems increasingly rely on Large Language Models (LLMs) yet the safety infrastructure surrounding these models introduces latency and compute overhead. This limits utility in resource-constrained, time-critical deployments. Existing external guardrail models remain blind to the model's internal workings, creating a fundamental assurance gap. We ask: does the model already know when the content is harmful? We extract activations from LLaMA-3.1-8B and train lightweight MLP classifier probes (12.6M parameters) to detect harmful prompts. Evaluated on WildJailbreak, Beavertails, and AEGIS 2.0, our probes achieve F1 scores of 99\%, 83\%, and 84\%, respectively competitive with 1000$\times$+ larger guard models while cutting latency and compute costs.
\end{abstract}


\section{Introduction}
\secgap
As LLMs are embedded into autonomous systems, from UAV mission planners \cite{xiao2025llm} and natural language command interfaces to space operations decision-support tools \cite{yuan2025next}, their capacity to produce harmful or misaligned outputs becomes a dependability concern, not merely a quality issue. Sometimes users deliberately try to elicit dangerous outputs \cite{niu2024jailbreaking} \cite{andriushchenko2024jailbreaking} \cite{chao2024jailbreakbench}; other times the model stumbles into them on its own \cite{zou2023universal, wei2024jailbroken}. Researchers have tackled this in two ways. \textit{Build-time} approaches like RLHF based alignment training \cite{ouyang2022training} bake safety into the model during training. These are brittle \cite{greenblatt2024alignment}. \textit{Run-time} approaches monitor the model during the inference loop.

The dominant run-time strategy uses external guardrails: separate models like LLaMA Guard \cite{inan2023llama}, ShieldGemma \cite{zeng2024shieldgemma}, and WildGuard \cite{han2024wildguard} that screen prompts and responses of the main LLM \cite{dong2024building}. But these guardrails have real drawbacks (Section~\ref{sec:guardrail_limitations}).

We ask: What if the LLM already knows when a prompt is harmful? Recent work suggests it does \cite{saglam2025large, chen2025towards, zou2023representation}. If that's true, we should be able to read that signal straight from the model's internals, minimizing the reliance on expensive external safety infrastructure.

We leverage the improved understanding of the geometry of an LLM's internal representations to train MLP probes on select activations from LLaMA-3.1-8B and evaluate on three safety benchmarks: WildJailbreak \cite{jiang2024wildjailbreak}, Beavertails \cite{ji2023beavertails}, and AEGIS 2.0 \cite{ghosh2025aegis}. Our results support the notion that the model's hidden states already encode useful safety signals which can be used to make safety determinations during inference.

\section{Related Work}
\label{sec:related}
\secgap
\subsection{External Guardrails}
\begin{figure}[H]
\centering
\includegraphics[width=\columnwidth]{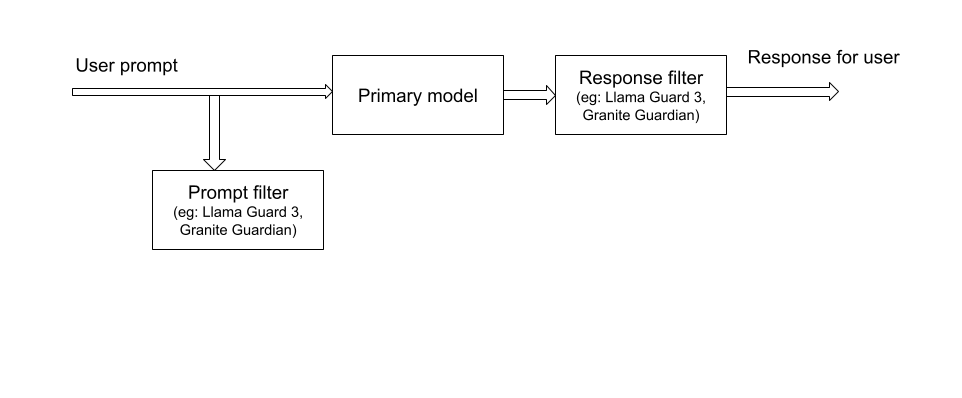}
\caption{External guardrail architecture. User prompts pass through a prompt filter on their way to the primary model, and responses pass through a response filter before reaching the user. Each filter requires a separate inference pass through the guard model.}
\label{fig:guardrails}
\end{figure}

Most production systems handle safety by adding separate guard models that check inputs and outputs \cite{dong2024building, inan2023llama, zeng2024shieldgemma}. Fig.~\ref{fig:guardrails} shows how this typically works.

This approach is the best one can do with black box access to the model. However, this has serious limitations (see Section~\ref{sec:guardrail_limitations}). Perhaps most importantly, external guards are blind to what happens inside the model, overlooking a very critical and information rich source of signal. Prompt injection works because injected instructions hijack the model's attention patterns \cite{hung2025attention} before any filter even sees the output. By the time an external filter checks the response, the attack has already succeeded, and the filter has limited visibility into \textit{what} happened \cite{zhang2024llmscan}.

\subsection{Safety Representations and Probes}
Recent interpretability research has isolated safety information within LLMs: \cite{zhang2024llmscan, chen2025towards} identified ``safety neurons,'' Zou et al.~\cite{zou2023representation} showed that safety concepts appear as linear directions in hidden states, and Saglam et al.~\cite{saglam2025large} achieved 96.7\% precision on jailbreak detection using MLP probes. Constitutional Classifiers++ \cite{cunningham2026constitutional} demonstrated that lightweight probes on activations can match external classifiers at 40$\times$ lower cost.

\subsection{Direct Threat Model: Adversarial Prompt Injection}
We focus on the adversarial prompt injection threat as the primary attack surface. In this threat model, an external attacker who has no access to model weights, training data, or system internals constructs malicious natural language inputs designed to cause the LLM to generate harmful, policy-violating, or operationally dangerous outputs. The attacker's goal is to bypass the model's alignment training, either by directly requesting harmful content or by embedding jailbreak constructions that reframe, roleplay, or obfuscate the harmful intent \cite{jiang2024wildjailbreak} \cite{andriushchenko2024jailbreaking} \cite{chao2024jailbreakbench}.

Formally, we define the threat as follows. Let $\mathcal{M}$ be an LLM deployed as a reasoning or command component in an autonomous system, and let $\mathcal{A}$ be an external adversary capable of injecting arbitrary text into $\mathcal{M}$'s input context. $\mathcal{A}$ seeks to find a prompt $p^{*} \in \mathcal{P}$ such that $\mathcal{M}(p^{*})$ produces output $o^{*}$ satisfying a harmful objective $\mathcal{H}(o^{*}) = 1$, where $\mathcal{H}$ is a harm classifier. The adversary is constrained to the input interface and has no knowledge of $\mathcal{M}$'s weights or intermediate activations, also referred to as a black-box attacker. This is the most operationally realistic threat model for deployed autonomous systems, where physical or network access to model internals is restricted.

The adversarial objective can be stated as:

\begin{equation}
    p^{*} = \argmax_{p \,\in\, \mathcal{P}_{\mathcal{A}}} \; \mathcal{H}\!\left(\mathcal{M}(p)\right)
    \quad \text{subject to} \quad \mathcal{G}\!\left(p,\, \mathcal{M}(p)\right) = 0
    \label{eq:threat_model}
\end{equation}

\noindent where $\mathcal{P}_{\mathcal{A}} \subseteq \mathcal{P}$ is the set of prompts accessible to the black-box adversary, $\mathcal{H} : \mathcal{O} \to \{0, 1\}$ is a harm classifier over the output space $\mathcal{O}$, and $\mathcal{G}$ represents the interface-level guardrail constraint that the adversary must satisfy to avoid pre-generation interception. The constraint $\mathcal{G}(\cdot) = 0$ captures the adversary's requirement to evade existing surface-level filters while still inducing a harmful output.

Existing external guardrail approaches attempt to intercept $p^{*}$ or $o^{*}$ at the interface boundary. As we have demonstrated, a sufficiently sophisticated $\mathcal{A}$ can construct inputs that satisfy surface-level safety filters while still inducing harmful internal model states. Our core claim is that we can detect harmful $p^{*}$  using the model's latent activations during processing of $p^{*}$ at a fraction of the cost of interface level filters leveraging a detection signal that is structurally inaccessible to the adversary.

\FloatBarrier
\section{Methodology}\label{sec:methodology}
%\vspace{-0.95\baselineskip}\noindent
We extract hidden states from LLaMA-3.1-8B and train MLP classifiers to predict whether prompts are harmful.

Algorithm~1 summarizes the two-stage training procedure. Stage 1 extracts activations: each prompt is passed through the primary LLM, and we store the final hidden state using last-token pooling. Stage 2 trains the classifier: the MLP learns to map stored activations to binary safety labels using cross-entropy loss.

\vspace{0.5em}
\noindent
\fbox{\parbox{\dimexpr\columnwidth-2\fboxsep-2\fboxrule\relax}{
\small
\textbf{Algorithm 1:} \textbf{Safety Probe Training}\\[0.3em]
\hrule
\vspace{0.3em}
\textbf{Input:} Dataset $\mathcal{D} = \{(x_i, y_i)\}$, LLM $\mathcal{M}$, epochs $E$, learning rate $\eta$\\
\textbf{Output:} Trained safety probe $f_\theta$\\[0.2em]
\begin{tabular}{@{}r@{\hspace{0.5em}}l@{}}
1: & \textbf{for} each prompt $x_i \in \mathcal{D}$ \textbf{do}\\
2: & \quad $h_i \leftarrow \mathcal{M}(x_i)[-1]$ \hfill\textit{// Last-token pooling}\\
3: & \quad Store $(h_i, y_i)$ where $h_i \in \mathbb{R}^{4096}$\\
4: & \textbf{end for}\\
5: & Initialize MLP $f_\theta: \mathbb{R}^{4096} \rightarrow \{0, 1\}$\\
6: & \textbf{for} epoch $= 1, \ldots, E$ \textbf{do}\\
7: & \quad \textbf{for} each batch $(H, Y)$ \textbf{do}\\
8: & \quad\quad $\hat{Y} \leftarrow f_\theta(H)$\\
9: & \quad\quad $\mathcal{L} \leftarrow \text{CrossEntropy}(\hat{Y}, Y)$\\
10: & \quad\quad $\theta \leftarrow \theta - \eta \nabla_\theta \mathcal{L}$\\
11: & \quad \textbf{end for}\\
12: & \textbf{end for}\\
13: & \textbf{return} $f_\theta$
\end{tabular}
}}

\subsection{System Architecture}

\begin{figure}[H]
\centering
\includegraphics[width=\columnwidth]{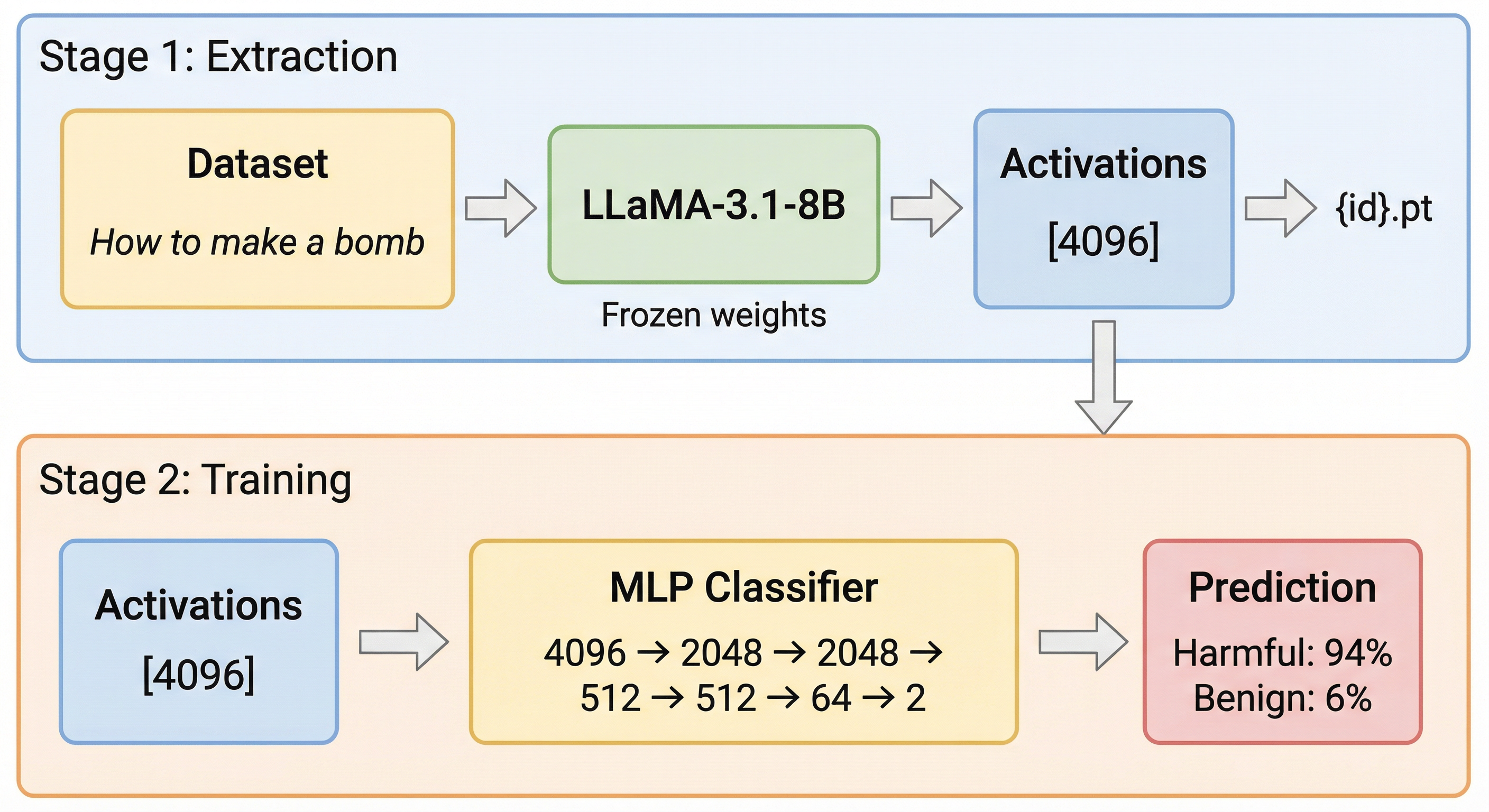}
\caption{Probe pipeline. Stage 1 extracts 4096-dim activations from final layer of frozen LLaMA-3.1-8B. Stage 2 trains a 6-layer MLP for binary classification.}
\label{fig:pipeline}
\end{figure}

Figure~\ref{fig:pipeline} shows the pipeline. Each prompt goes through LLaMA-3.1-8B, and we pull out the 4096-dimensional activation from the final layer using last-token pooling. From there, a 6-layer MLP classifies the embedding as harmful or benign.

\begin{figure}[H]
\centering
\includegraphics[width=\columnwidth]{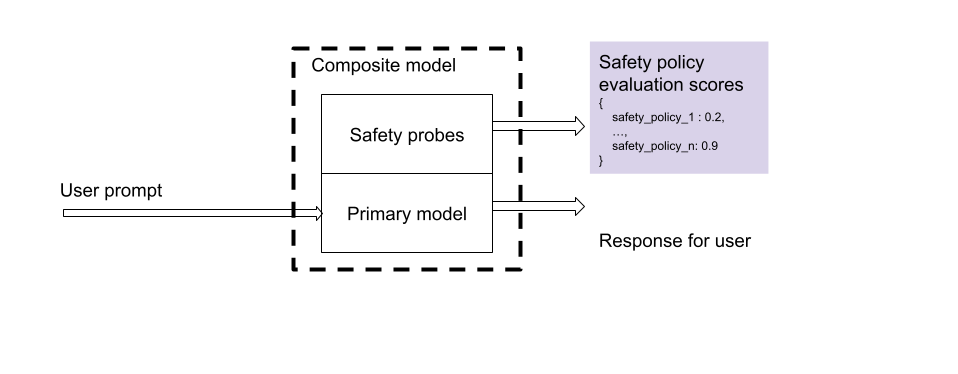}
\caption{Deployment architecture. Safety probes run alongside the primary model, producing per-policy risk scores without external infrastructure. Compare with external guardrails (Fig.~\ref{fig:guardrails}).}
\label{fig:probes}
\end{figure}

\subsection{Dataset Preprocessing}

We evaluated our probes on three separate safety datasets, each requiring different preprocessing for binary classification.

\textbf{WildJailbreak} \cite{jiang2024wildjailbreak} provides 261K prompts spanning four categories: vanilla harmful, vanilla benign, adversarial harmful, and adversarial benign. We focus on the vanilla subset (roughly 100K examples) because our goal is classifying prompt intent, not detecting adversarial jailbreak attacks. We label harmful prompts as 1 and benign as 0, then split 80/10/10 for training, validation, and testing, stratified by label.

\textbf{Beavertails} \cite{ji2023beavertails} contains $\sim$364K English question-answer pairs labeled with 14 harm categories (e.g., violence, hate speech, discrimination) and a binary \texttt{is\_safe} flag. The dataset includes structured train/test splits and a curated evaluation set of $\sim$700 prompts for benchmarking.

\textbf{AEGIS 2.0} \cite{ghosh2025aegis} includes 26K human-LLM interactions labeled across 12 hazard categories such as violence, hate speech, and self-harm. Some sensitive prompts are redacted in the public release but can be reconstructed using IDs linked to the Kaggle Suicide Watch dataset; we reconstruct these to ensure complete coverage. Unsafe prompts are labeled 1, safe prompts 0, using the provided 80/10/10 train/validation/test splits.

\subsection{Classifier Architecture}

\begin{table}[H]
\centering
\caption{MLP architecture (12.6M parameters, 0.16\% of LLaMA-8B).}
\label{tab:architecture}
\begin{tabular}{lccc}
\toprule
\textbf{Layer} & \textbf{In} & \textbf{Out} & \textbf{Activation} \\
\midrule
1 & 4,096 & 2,048 & GELU + Dropout \\
2 & 2,048 & 2,048 & GELU + Dropout \\
3 & 2,048 & 512 & GELU + Dropout \\
4 & 512 & 512 & GELU + Dropout \\
5 & 512 & 64 & GELU + Dropout \\
6 & 64 & 2 & Softmax \\
\bottomrule
\end{tabular}
\end{table}

Following Saglam et al.~\cite{saglam2025large}, we use a 6-layer MLP with progressively decreasing dimensions (Table~\ref{tab:architecture}). The classifier contains 12.6M parameters, only 0.16\% of LLaMA's 8B. We train with AdamW \cite{loshchilov2019decoupled}, learning rate $2.5 \times 10^{-4}$, batch size 1024, for 50 epochs.

\section{Results}

\begin{table}[H]
\centering
\caption{Probe performance across datasets.}
\label{tab:performance}
\begin{tabular}{lccc}
\toprule
\textbf{Dataset} & \textbf{Precision} & \textbf{Recall} & \textbf{F1 (\%)} \\
\midrule
WildJailbreak & 0.997 & 0.985 & 99.1 \\
Beavertails & 0.841 & 0.814 & 82.7 \\
AEGIS & 0.858 & 0.814 & 83.5 \\
\bottomrule
\end{tabular}
\end{table}

Table~\ref{tab:performance} shows the performance of our probes on LLaMA-3.1-8B. We achieve 99.1\% F1 on WildJailbreak, 82.7\% F1 on Beavertails and 83.5\% F1 on AEGIS datasets. The precision and recall numbers for each one of these datasets are also within a 5\% range of each other, highlighting stable detection performance.

\begin{table}[H]
\centering
\small
\caption{Comparison with guard models. BeaverTails results from Han et al.~\cite{han2024wildguard}; AEGIS 2.0 results from Ghosh et al.~\cite{ghosh2025aegis}.}
\label{tab:comparison}
\begin{tabular}{lccc}
\toprule
\textbf{Model} & \textbf{Params} & \textbf{BeaverT} & \textbf{AEGIS 2.0} \\
\midrule
BeaverDam & 7B & 89.9 & -- \\
MD-Judge & 7B & 86.7 & -- \\
GPT-4 & $\sim$1.8T & 86.1 & -- \\
WildGuard & 7B & 84.4 & 81.9 \\
LlamaGuard3 & 8B & -- & 77.3 \\
LlamaGuard2 & 8B & 71.8 & 76.8 \\
Aegis-Guard & 7B & 74.7 & -- \\
\midrule
\textbf{Our Probe} & 12.6M & 82.7 & 83.5 \\
\bottomrule
\end{tabular}
\end{table}

Table~\ref{tab:comparison} compares our probe against established guard models. On BeaverTails, our 12.6M-parameter probe achieves 82.7\% F1, competitive with 7B models like WildGuard (84.4\%) and MD-Judge (86.7\%), while substantially outperforming LlamaGuard2 (71.8\%). On AEGIS 2.0, we achieve 83.5\% F1, surpassing WildGuard (81.9\%), LlamaGuard3 (77.3\%), and LlamaGuard2 (76.8\%). Top performers like BeaverDam (89.9\%) and GPT-4 (86.1\%) score higher on BeaverTails but they are also over 1000$\times$ larger. We note that BeaverTails evaluates response harmfulness while AEGIS 2.0 evaluates prompt classification. We omit WildJailbreak because Han et al.~\cite{han2024wildguard} and others evaluate on WildGuardTest, a different test set than the vanilla split we use.

\section{Discussion}
\secgap

Our results demonstrate that lightweight probes on LLM hidden states can detect harmful content with high accuracy across diverse safety benchmarks. To contextualize these findings, we first examine the limitations of external guardrails, then discuss how probe-based approaches address them.

\subsection{Limitations of External Guardrails}
\label{sec:guardrail_limitations}

\textbf{Latency and Cost.} Guardrails require dedicated GPUs and add latency to every request. Total latency includes both compute and network overhead:
\begin{equation}
\label{eq:guard}
L_{\text{guard}} = T_{\text{pf}} + T_{\text{llm}} + T_{\text{rf}} + 2T_{\text{net}}
\end{equation}
where $T_{\text{pf}}$, $T_{\text{llm}}$, $T_{\text{rf}}$ are the response times of the prompt filter, the LLM, and the response filter respectively.
These operations introduce sequential dependencies: the LLM must finish before the response filter can evaluate its output, and both prompt and response filters must finish running before the result can be sent to the user. Each filter also incurs network latency when deployed as an external service. Compute overhead alone can add significant latency. Albrethsen et al.~\cite{albrethsen2026deepcontext} report latency numbers between 4--1430~ms per turn for different guard models. The network latencies are implementation specific and grow with payload size. The latency overhead has motivated a growing line of research on shrinking guard models \cite{zheng2024lightweight, fedorov2024llamaguard}. Zheng et al.~\cite{zheng2024lightweight} highlight an approach to shrink the guard model from 7B to 67M parameters to achieve acceptable latency while trading off on safety.

\textbf{Adversarial Vulnerability.} Guardrails remain vulnerable despite years of development. Adversarial techniques like character injection achieve up to 100\% evasion rates against commercial systems including Microsoft Azure Prompt Shield and Meta Prompt Guard \cite{hackett2025bypassing, yang2025guiding, geng2025safety, fairoze2025bypassing, reddy2025autoadv}.

\textbf{Opacity to Internal Mechanisms.} Guard models see inputs and outputs, nothing in between. During a Prompt injection attack, injected instructions hijack the model's internal representations including the attention layer. \cite{hung2025attention}, and hidden states encode malicious intent before any tokens are generated \cite{chia2025probing, saglam2025large}. By the time an external filter sees the response, the attack has already succeeded. 

This creates a fundamental asymmetry of information. The attacker operates entirely in the input space — crafting prompts that satisfy surface-level filters while steering internal representations toward harmful generation — and never needs to observe the model's internals to succeed. The defender, relying solely on external guardrails, is constrained to the same input-output interface as the attacker and so holds no informational advantage.

\subsection{Advantages of Probe-Based Detection}

Our approach (Fig.~\ref{fig:probes}) addresses each of these limitations directly.

\textbf{Minimal Latency.} Probes integrate into the forward pass rather than running separately:
\begin{equation}
\label{eq:probe}
L_{\text{probe}} = \max(T_{\text{llm}}, T_{\text{probe}})
\end{equation}
While guardrail latency is additive (\eqref{eq:guard}), probe latency uses $\max$ because the probe executes concurrently with generation on the same host. Since the number of probe parameters (12.6M) $\ll$ number of primary model parameters (8B), it follows that $T_{\text{probe}} \ll T_{\text{llm}}$. The probe is effectively ``free.'' In practice, we observed that the probes ran in under 1~ms while typical generation took 50--500~ms. Our probes do not require separate GPUs for running guard models and eliminate network transport costs entirely.

\textbf{Early Termination.} Unlike external guardrails that must wait for complete responses, probes can flag harmful content during generation and terminate immediately, saving compute and limiting user exposure to unsafe content.

\textbf{Access to Internal State} Probes operate on hidden representations where harmful intent is encoded. The defender gains access to the model's internal activations, a signal that is both richer than the input-output interface and structurally inaccessible to a black-box adversary. Under this information symmetry driven framing, probing hidden states is not merely an efficiency improvement over external guardrails — it is a shift in the defender's observability that the attacker cannot trivially counter without white-box access to the model.

\section{Conclusion}
\secgap

We showed that lightweight probes trained on frozen LLM activations detect harmful content effectively. On three benchmarks, our approach achieved F1 scores of 99\% (WildJailbreak), 83\% (Beavertails), and 84\% (AEGIS 2.0). These probes have 12.6M parameters, less than 0.2\% of the base model, and need no external infrastructure.

The bottom line is that safety-relevant information already lives in LLM hidden states, organized in linearly separable form. Simple classifiers can tap this structure directly, providing a fast and lightweight alternative to external guardrails by re-using safety signals within the primary model.

\section*{Limitations}
\secgap

Our experiments focus exclusively on LLaMA-3.1-8B; we have not validated cross-model transfer. We extract only final-layer activations. The earlier layers may encode safety information differently. Our probes predict binary labels; real deployments may require finer-grained risk scores.

\textbf{Future directions} include analyzing activations at different layers and tokens to trace how harmful intent emerges, evaluating adversarial robustness, and expanding to other model architectures and modalities.

\section*{Impact Statement}
\secgap

Large language models are increasingly deployed in real-world applications where failures can produce harmful or policy-violating outputs. This work introduces latent space probes, a class of model-native runtime defenses that analyze internal model activations to detect safety-relevant concepts during generation. By operating directly on internal representations, such defenses may enable faster and more scalable detection of unsafe behaviors, including prompt injection attacks and harmful content, while avoiding the latency and computational costs of external moderation systems. This approach could help improve the reliability and safety of AI systems deployed in production.

However, activation-based detection systems may inherit biases present in underlying models or datasets, which could lead to disproportionate moderation of certain linguistic styles or topics. In addition, such mechanisms could be misused for overly restrictive filtering if deployed without transparency or appropriate governance. Careful evaluation across diverse datasets and responsible deployment practices are therefore important to mitigate these risks.

\bibliographystyle{IEEEtran}
\bibliography{references}

\end{document}